\documentclass[letterpaper, 10 pt, conference]{ieeeconf}  %

\IEEEoverridecommandlockouts                              %

\usepackage{amsmath,amsfonts}
\usepackage{algorithmic}
\usepackage{algorithm}
\usepackage{array}
\usepackage{hyperref}
\usepackage{cleveref}
\usepackage[caption=false,font=normalsize,labelfont=sf,textfont=sf]{subfig}
\usepackage{textcomp}
\usepackage{stfloats}
\usepackage{url}
\usepackage{verbatim}
\usepackage{graphicx}
\usepackage{cite}
\usepackage{lineno}
\usepackage{xcolor}
\usepackage{xspace}
\usepackage{adjustbox}
\usepackage{booktabs}
\usepackage{multirow}
\usepackage{scalerel}
\usepackage{dcolumn}
\usepackage{todonotes}
\usepackage{pdfcomment}
\newcolumntype{/}{D{/}{/}{-1}}
\def \MethodAcronym {CorDex\xspace}

\title{\LARGE \bf
Generate, Transfer, Adapt: Learning Functional Dexterous Grasping from a Single Human Demonstration
}

\author{Xingyi He$^{*}$, Adhitya Polavaram, Yunhao Cao, Om Deshmukh, Tianrui Wang, Xiaowei Zhou, Kuan Fang$^{\dagger}$
\thanks{$^{*}$ This work was conducted while Xingyi He was a visiting scholar at Cornell University.}%
\thanks{$^{\dagger}$ Corresponding Author}%
}

\begin{document}

\maketitle
\thispagestyle{empty}
\pagestyle{empty}

\begin{abstract}

Functional grasping with dexterous robotic hands is a key capability for enabling tool use and complex manipulation, yet progress has been constrained by two persistent bottlenecks: the scarcity of large-scale datasets and the absence of integrated semantic and geometric reasoning in learned models. In this work, we present \MethodAcronym, a framework that robustly learns dexterous functional grasps of novel objects from synthetic data generated from just a single human demonstration. At the core of our approach is a correspondence-based data engine that generates diverse, high-quality training data in simulation. Based on the human demonstration, our data engine generates diverse object instances of the same category, transfers the expert grasp to the generated objects through correspondence estimation, and adapts the grasp through optimization. Building on the generated data, we introduce a multimodal prediction network that integrates visual and geometric information. By devising a local–global fusion module and an importance-aware sampling mechanism, we enable robust and computationally efficient prediction of functional dexterous grasps. Through extensive experiments across various object categories, we demonstrate that \MethodAcronym generalizes well to unseen object instances and significantly outperforms state-of-the-art baselines.
For additional results and videos, please visit \url{https://cordex-manipulation.github.io}.

\end{abstract}

\section{Introduction}

\begin{figure}[t]
    \centering
    \includegraphics[width=1.0\linewidth]{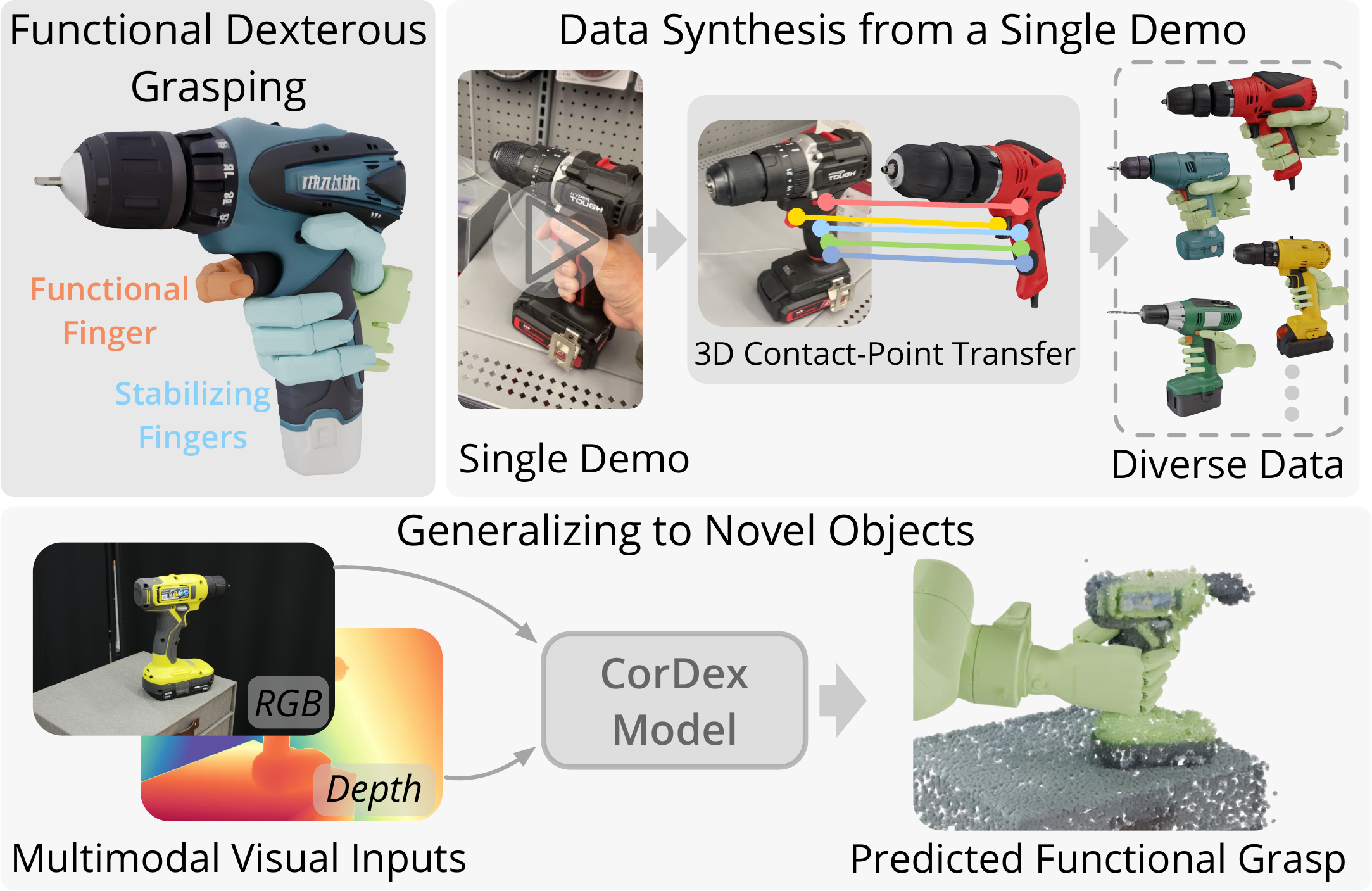}
    \vspace{-0.7cm}
    \caption{
    \MethodAcronym learns to robustly perform functional dexterous grasping by combining a correspondence-based data engine and a multimodal grasp prediction model. The data engine scales a single human demonstration into diverse high-quality grasp data on novel objects. By learning from the generated data, the \MethodAcronym model leverages multimodal inputs to predict grasps for novel object instances.
    }
    \vspace{-0.4cm}
    \label{fig:teaser}
\end{figure}

Functional grasping with dexterous hands is a fundamental capability that enables robots to perform complex tool use and precise manipulation.
Unlike conventional grasping with simple end-effectors~\cite{Fang2022AnyGraspRA} or task-agnostic methods focused solely on stability~\cite{Wang2022DexGraspNetAL,Wei2024DROG}, dexterous functional grasping requires predicting high-dimensional motor commands that jointly satisfy both physical and semantic constraints~\cite{Li1987TaskorientedOG, Agarwal2023DexterousFG}.
In particular, the robot must not only establish a stable hold on the object but also meaningfully interact with its task-relevant part in order to realize its intended functionality, as illustrated in Fig.~\ref{fig:teaser}. 
Satisfying these demands under contact-rich interactions and fine-grained control makes dexterous functional grasping a persistent challenge.

An increasing number of recent works have explored functional dexterous grasping with learning-based approaches. However, despite encouraging progress, advancement remains constrained by two fundamental bottlenecks. First, acquiring large-scale, high-quality datasets with functional dexterous grasp annotations is prohibitively difficult. Real-world data collection through motion capture or teleoperation~\cite{Brahmbhatt2019ContactDBAA,liu2024realdex,Yang2022OakInkAL} demands extensive human effort and scales poorly to novel objects and tasks. Alternatively, methods that leverage in-the-wild human video demonstrations~\cite{Kannan2023DEFTDF,Chen2025Web2GraspLF} offer broader coverage but suffer from severe reconstruction and pose estimation errors, necessitating costly data curation. Second, even with sufficient data, most approaches focus on geometric reasoning over object shape, which is inadequate for capturing semantic cues about functionality. Without jointly exploiting semantic and geometric information, these models often fail to produce grasps that are both physically stable and functionally appropriate in unseen scenarios.

In this paper, we present \MethodAcronym, a framework that enables learning robust dexterous functional grasping from a single human demonstration video. \MethodAcronym combines a data engine that autonomously produces diverse, high-quality training data in simulation with a novel prediction model that effectively integrates visual and geometric information to compute functional dexterous grasps for novel objects.

We devise a three-stage data engine to scale up functional dexterous grasping data from a single human video demonstration. First, given the object category in the video, the data engine generates diverse object instances by retrieving Internet images and converting them into 3D models. Second, the expert grasp from the demonstration is transferred to each generated object using a novel correspondence-based pipeline. Finally, to ensure label quality, we introduce a physics-informed adaptation procedure that optimizes the transferred grasps in simulation.
Unlike prior work that relies on 3D correspondence estimation~\cite{Wang2023SparseDFFSF,Zhu2024DenseMatcherL3}, which performs poorly due to the significant appearance and shape gap across different instances, our pipeline produces diverse and high-quality training data for functional dexterous grasping.

Building on the generated grasping data, we propose a prediction model, which learns to infer dexterous functional grasps from single-view RGB-D input. In contrast to previous approaches that rely solely on object geometry~\cite{Wei2024DROG,liu2024realdex}, our model jointly reasons over semantic cues from images and geometric properties from point clouds. To achieve this, we design a local-global fusion module that integrates features from both modalities. In addition, we introduce a sampling mechanism that adaptively focuses on regions where robot-object interactions are likely to occur, improving both computational efficiency and prediction accuracy. Together, these components enable our model to produce grasps that are not only physically robust but also functionally meaningful for novel objects.

The proposed functional grasping data engine autonomously generates 11 million grasp-image pairs for 900 diverse objects across nine categories with minimal human annotation effort.
Using both the generated diverse functional grasping data and the effective grasp prediction network, CorDex robustly generalizes to unseen objects from single-view input.
We validate our approach through extensive experiments in simulation and the real world, spanning nine object categories and two robot embodiments. On unseen real-world objects, our method achieves a $69$\% success rate, substantially outperforming state-of-the-art methods.

\section{Related Works}

\noindent \textbf{Dexterous grasping.}
Grasping has long been a fundamental task in robotics. Early methods for dexterous hands planned grasps using analytic metrics such as force-closure and wrench-based quality measures~\cite{Ferrari1992PlanningOG,Borst2004GraspPH,Lin2015GraspPT}. With the rise of deep learning, data-driven approaches emerged that predict grasp configurations in different forms, including joint angles~\cite{Xu2023UniDexGraspUR,Geng2023UniDexGraspID,pi2025coda,zhang2025RobustDexGrasp}, contact regions~\cite{Shao2019UniGraspLA,Li2022GenDexGraspGD}, and distance matrices between hand and object points~\cite{Wei2024DROG}. These methods typically utilize geometry information derived from complete meshes~\cite{Shao2019UniGraspLA,Li2022GenDexGraspGD} or partial point clouds~\cite{Wei2024DROG,zhang2025RobustDexGrasp}. To construct training data, grasp generation is often formulated as an optimization problem satisfying stability constraints~\cite{Wang2022DexGraspNetAL,Chen2025DexonomySA}. While effective for stable grasping, these geometry-centric approaches neglect semantic cues critical for functionality. 
Our work differs by jointly leveraging semantic features from RGB images and geometric features from point clouds, enabling predictions that are both stable and functional.
Grasping has also been studied through category-level pose estimation, which predicts the 6D pose and 3D size of unseen objects from RGB-D data using category-level pretraining~\cite{Wang2019NormalizedOC,Lin2024InstanceAdaptiveAG}, and transfers grasps from a reference object via the estimated transformation. However, because pose estimation provides only coarse alignment and ignores fine-grained shape variations, the transferred grasps often miss the correct functional regions. In contrast, our approach directly predicts grasp gestures, enabling flexible generalization to unseen objects with large shape variations.

\vspace{1mm}
\noindent \textbf{Functional grasping and tool use.}
Functional grasping requires contact with task-relevant object regions that afford intended use while ensuring stability. Classical approaches introduced task-oriented grasp metrics for dexterous hands~\cite{Li1987TaskorientedOG,Haschke2005TaskorientedQM} and explored affordance reasoning or keypoint prediction for parallel-jaw grippers~\cite{Zhu2015UnderstandingTT,Fang2018LearningTG}. More recent work has pursued learning-based solutions by collecting functional dexterous grasp data via motion capture~\cite{Yang2022OakInkAL}, teleoperation~\cite{liu2024realdex}, retargeting~\cite{Huang2025FunGraspFG}, or multimodal sensing~\cite{Brahmbhatt2019ContactDBAA}. However, these approaches require extensive human effort and scale poorly. To improve scalability, Internet demonstrations have been leveraged~\cite{Kannan2023DEFTDF,Chen2025Web2GraspLF}, though reconstruction errors limit label quality. 
Some one-shot approaches generate functional grasps by transferring contact information across instances through dense 3D correspondences~\cite{Wu2023FunctionalGT,Wei2024LearningHF,Wang2023SparseDFFSF,Zhu2024DenseMatcherL3}.
However, these methods fall short due to limited generalizability to unseen objects, even within the same category, due to the limited 3D correspondence training data.
In contrast, we propose a correspondence-based data engine that robustly transfers contact information across categories with a multimodal prediction model that fuses semantic and geometric information from RGB-D input to achieve robust prediction and category-level generalization to novel objects based on only a single human demonstration video for each object category.  

\vspace{1mm}
\noindent \textbf{Robot learning from synthetic data.}
Synthetic data has become a powerful enabler for scalable robot learning, reducing dependence on costly real-world annotation~\cite{tobin2017domainrandomization,peng2018simtoreal, fang2023active, Mu_2025_CVPR, lin2024twisting, wang2023dexgraspnetlargescaleroboticdexterous}. Prior works augment real-world demonstrations by generating diverse variants in simulation~\cite{jiang2024dexmimicen,dai2024acdc, maddukuri2025simandreal}. Recent advances in image matching~\cite{Caron2021EmergingPI,Oquab2023DINOv2LR,wang2024eloftr,he2025matchanything} further enable scalable annotation transfer across object instances. Building on these ideas, our contribution targets the unique challenges of dexterous functional grasping. We introduce a correspondence-based engine that generates diverse, contact-rich grasps from a single human video demonstration.

\begin{figure*}[t]
    \centering
    \vspace{-0.35cm}
    \includegraphics[width=1.0\linewidth]{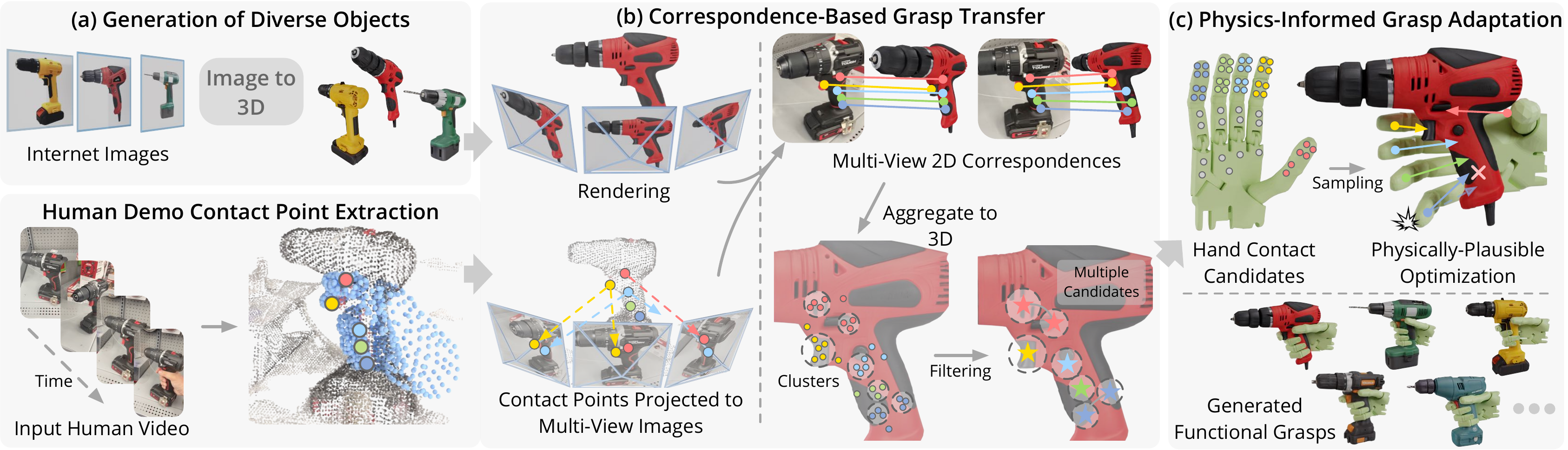}
    \vspace{-0.8cm}
    \caption{
    \textbf{\MethodAcronym data engine.} 
    We generate diverse, high-quality functional grasps for novel objects from a single human demonstration through three stages: 
    (a) \textit{Generate}: diversify objects within the task category by creating 3D models from Internet-retrieved images. 
    (b) \textit{Transfer}: extract 3D fingertip contacts~(\protect\scalerel*{\includegraphics{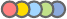}}{X}) from the demonstration via scene and hand reconstruction, then transfer them to novel objects using a correspondence-based 2D–3D pipeline that projects, matches, and aggregates contact points into reliable 3D candidates~(\protect\scalerel*{\includegraphics{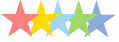}}{\strut}) on generated objects. 
    (c) \textit{Adapt}: apply physics-informed grasp adaptation to convert candidate contact points into embodiment-specific grasps that satisfy both functionality and stability considerations, yielding diverse and high-quality functional grasp data.
    }
    \vspace{-0.5cm}
    \label{fig:dataengine}
\end{figure*}

\section{Preliminaries}
\label{sec:preliminaries}

\textbf{Dexterous functional grasping.}  
Extending the problem formulation from \cite{Kannan2023DEFTDF, Chen2025Web2GraspLF, Agarwal2023DexterousFG}, we consider a robotic hand with $M$ fingers and $K$ degrees of freedom. A dexterous grasp is defined as $g = (T, \theta)$, where $T \in SE(3)$ is the hand pose and $\theta \in \mathbb{R}^K$ represents finger joint angles. We require each grasp to satisfy two criteria: \emph{functionality} and \emph{stability}. A grasp is functionally appropriate if a designated set of fingers $\mathcal{F} \subseteq \{1, \dots, M\}$ establishes contact with the corresponding functional regions $\mathcal{R}_f \subseteq \mathbb{R}^3$ 
of the object:
\begin{equation}
\forall f \in \mathcal{F}, \;\; \exists p_f \in \mathcal{R}_f\;\; \text{s.t.} \;\; \text{dist}\big(h_f(g), p_f\big) < \epsilon,
\end{equation}
where $h_f(g)$ is the fingertip position of finger $f$ and $\epsilon > 0$ is a tolerance. 
A grasp is defined as stable if a subset of stabilizing fingers $\mathcal{S} \subseteq \{1, \dots, M\}$ can securely hold the object and resist external forces acting on the object, thereby maintaining its pose relative to the hand under perturbations.
Together, these criteria capture the dual requirements that the grasp aligns with the object’s intended functionality while being robust to external disturbances. 
The robot receives a single-view RGB-D image and the target object mask, both aligned to the base frame through hand–eye calibration, and predicts functional grasps.

\textbf{Grasp prediction with $\mathcal{D(R,O)}$.}
To efficiently predict high-dimensional dexterous grasps, we build upon the $\mathcal{D(R,O)}$ paradigm\cite{Wei2024DROG}, which uses a \emph{dense distance matrix} between points sampled on the robot hand and object to represent a grasp. From this representation, $g$ is recovered by multilateration~\cite{norrdine2012algebraic} and inverse kinematics. Such a representation removes the need for expensive collision terms during optimization, and naturally generalizes across different hand embodiments.
$\mathcal{D(R,O)}$ additionally formulates the policy network as a conditional variational autoencoder (CVAE), allowing the model to capture the multimodality of grasps contained in the training dataset. During training, the ground-truth hand configuration is encoded into a latent vector, which is concatenated with the fused object features to condition the distance matrix decoder. At test time, diverse grasps can be generated by discarding the latent encoder and directly sampling the latent from a prior distribution. The training objective combines three components: (i) an L1 loss between the predicted and ground-truth distance matrices, (ii) KL divergence regularization on the CVAE latent space, and (iii) a pose error loss supervising the reconstructed grasp. 
Building on the $\mathcal{D(R,O)}$ framework, our model adopts the dense distance matrix representation and the CVAE formulation, while introducing several key enhancements to enable accurate functional grasping.

\section{Correspondence-Based Data Synthesis}
\label{subsec:dataengine}

To enable scalable learning of functional dexterous grasping, we generate synthetic grasp data for each object category from a single human demonstration video. Such demonstrations can be easily captured with affordable devices such as smartphone cameras, avoiding the need for expensive multi-camera or teleoperation setups and allowing functional grasp annotations with minimal manual effort. The central challenge, however, is how to scale from a single demonstration to diverse objects while ensuring that the synthesized grasp labels remain consistent with the demonstration video.

To address this challenge, we introduce a data engine that operates in three stages. First, we generate diverse object models to support grasp synthesis~(Fig.~\ref{fig:dataengine}a). Next, we extract 3D contact points from a single human demonstration to represent hand–object interactions, and transfer these points to novel objects through a robust cross-instance 2D–3D correspondence pipeline~(Fig.~\ref{fig:dataengine}b), which leverages advances in image matching and geometric cues to mitigate noise and inconsistencies. Finally, we apply a physics-informed grasp adaptation procedure that optimizes robot hand configurations with respect to the transferred contacts, ensuring both functional alignment and physical stability~(Fig.~\ref{fig:dataengine}c). The following subsections detail the design of each stage.

\subsection{Generation of Diverse Objects}

To enable category-level generalization, we generate diverse 3D object models that capture large intra-class variations while preserving the functional semantics of the task.
Instead of relying on fixed 3D datasets with limited coverage or text-to-3D generation, where vague descriptions often lead to unrealistic shapes, we adopt a 2D-to-3D generation approach that leverages the broad visual diversity of Internet images.
Starting from the demonstration video, we retrieve a large collection of Internet images of the same object category. 
Candidate images are filtered using pretrained visual feature similarity~\cite{Oquab2023DINOv2LR} to the demonstration, ensuring both diversity and relevance.
In cases where the task category lacks sufficient Internet images, we augment them using GPT-Image~\cite{openai_gpt_image_2025} inpainting, and generate high-quality 3D object models with Rodin~\cite{hyper3d_ai_bibtex}. The retrieved images are then used as conditions for a 2D-to-3D generation model~\cite{hyper3d_ai_bibtex}, producing massive, high-quality object meshes per object category. Compared to using a fixed object dataset, this approach creates realistic and diverse assets tailored to the demonstration, providing a strong foundation for cross-instance functional grasp transfer.

\subsection{Correspondence-Based Grasp Transfer}

With diverse object models generated for each task category, the next step is to transfer functional grasp knowledge from the human demonstration onto these novel instances. Directly retargeting human hand poses to a robot is infeasible due to object misalignment and the morphology gaps between human and robot hands.
Instead, we represent human–object interaction through \emph{3D fingertip contact keypoints}, which are embodiment-agnostic and transferable across objects. From the demonstration video, we reconstruct the hand mesh~\cite{Potamias2024WiLoRE3} and the object point cloud~\cite{Wang2025VGGTVG}, and then extract fingertip contacts as the nearest object surface points.
Since the reconstructed point clouds lack absolute scale, we determine the optimal scale by aligning the object to the hand mesh and minimizing the distances between fingertip points and their nearest object points.

Transferring these contact points to diversified objects is challenging because of large appearance and geometry variations. Naively applying cross-instance \emph{3D} matching~\cite{Zhu2024DenseMatcherL3,Wang2023SparseDFFSF} performs poorly (Sec.~\ref{sec:experiment}) due to limited training data and weak generalization. To overcome this, we leverage large-scale pretrained \emph{2D} matching models~\cite{Oquab2023DINOv2LR,he2025matchanything}, which generalize across categories, and couple them with a robust 3D aggregation step.
Specifically, the fingertip contacts are projected onto all valid frames of the demonstration, while novel objects are rendered from viewpoints uniformly sampled on a sphere.
A 2D matcher~\cite{he2025matchanything} establishes correspondences between demonstration frames and rendered images, enabling contact points to be transferred to novel object renderings and subsequently back-projected into 3D using the known camera intrinsics, extrinsics, and depth.

Because 2D matching can be noisy and view-inconsistent, the back-projected points from multiple views are aggregated in 3D with density-based clustering. We retain the centers of the three largest clusters as candidate contact locations for each fingertip and discard smaller clusters as outliers.
To further improve reliability, each candidate is weighted by the average 2D matching confidence of its member points.
The resulting candidate set provides multiple plausible, confidence-weighted hypotheses for each fingertip, which are then resolved through the physics-informed grasp optimization. Preserving this set of hypotheses explicitly models cross-instance ambiguity and allows downstream optimization to exploit geometric and physical constraints to select stable, functional grasps.

\subsection{Physics-Informed Grasp Adaptation}
Based on the candidate contact points on novel objects that specify the target locations for each finger, our goal is to generate embodiment-specific grasp labels for downstream model training.
However, variations in object scale between the demonstration and generated objects, together with correspondence noise, may make the transferred contact points unreachable for the robot hand.
To address this, we introduce a grasp adaptation process that jointly optimizes contact-point alignment and physical plausibility, ensuring that the resulting grasps are both functional and stable.

Specifically, for a robot hand, a set of candidate contact points is defined on every finger, which are likely to correspond to the transferred contact points on the object.
Considering size variations between the robot and human hands, we define candidate contact points on both the middle and distal links to provide more flexibility in aligning with the transferred contacts, as shown in Fig.~\ref{fig:dataengine}c.
Our pipeline simultaneously initializes $N$ grasps $g$ and optimizes them with hand contact points and object contact points sampled from the candidates.
The optimization objective is composed of the following terms:
\begin{itemize}
    \item \textit{Contact-prior loss} encourages sampled points on the hand surface to align with sampled contact points on the object surface, minimizing both positional distance and the deviation between normals:  
    \begin{equation}
    \resizebox{0.8\hsize}{!}{$
        \mathcal{L}_{prior} = 
        \sum_{l \in \mathcal{C}}
        \big( \| h_l(g) - o_p \|_2^2
        + \alpha \big(1 - n_{h}^\top n_{o}\big) \big),
    $}
    \end{equation}
    where $\mathcal{C}$ denotes the set of finger links with defined contact points, $h_l(g)$ is the 3D position of a sampled finger contact point on link $l$ under grasp $g$, $o_p$ is the transferred prior contact point on the object surface, $n_{h}$ and $n_{o}$ are their surface normals, and $\alpha$ is a hyper-parameter balancing positional and orientation terms.

    \item \textit{Stability contact loss} addresses scale mismatch (transferred contact points may be too dense or sparse to be reachable).
    This loss aligns sampled finger points (the same points as those used in the \textit{contact-prior loss}) with their nearest object surface samples, reducing unstable floating gestures caused by object scale misalignment:
    \begin{equation}
        \mathcal{L}_{stab} =
        \sum_{l \in \mathcal{C}}
        \| h_l(g) - o_c \|_2^2,
    \end{equation}
    where $h_l(g)$ is a sampled finger contact point under hand configuration $g$, and $o_c$ denotes its closest neighbor on the object surface.

    \item \textit{Auxiliary contact loss.} Since the transferred points constrain only the middle and distal finger links, we additionally sample points on other hand links (e.g., the palm) and encourage them to contact the object surface, thereby improving overall stability:
    \begin{equation}
        \mathcal{L}_{aux} =
        \sum_{l \in \mathcal{A}}
        \| h_l(g) - o_c \|_2^2,
    \end{equation}
    where $\mathcal{A}$ is the set of auxiliary hand links, $h_l(g)$ is a sampled contact point on link $l$, and $o_c$ denotes its closest neighbor on the object surface.

    \item \textit{Joint limit loss} penalizes violations of joint angle limits to ensure feasible configurations.
    \item \textit{Collision loss} penalizes the robot-object penetration.

    \item \textit{Self-penetration loss} penalizes robot links' penetration.
\end{itemize}

The formulations for \textit{joint-limit}, \textit{collision}, and \textit{self-penetration losses} follow~\cite{Wang2022DexGraspNetAL}.
The final objective is the weighted sum of all loss terms and is optimized using the method in~\cite{Liu2021SynthesizingDA}. After optimization, we obtain candidate grasps, which are verified in simulation to ensure physical stability~\cite{Wang2022DexGraspNetAL}. Specifically, each grasp is tested in a real-time physics engine~\cite{Makoviychuk2021IsaacGH} by applying external forces from six directions to check whether the object remains securely held. Verified grasps are retained as functional grasping data for training the prediction network. 
Finally, to produce large-scale training data for grasp prediction, we render photo-realistic RGB-D images in Blender~\cite{blender_homepage}, placing objects with random poses within a $1\text{m}^3$ cube and enhancing diversity with randomized backgrounds and lighting.

\begin{figure*}[t]
    \centering
    \vspace{-0.45cm}
    \includegraphics[width=1.0\linewidth]{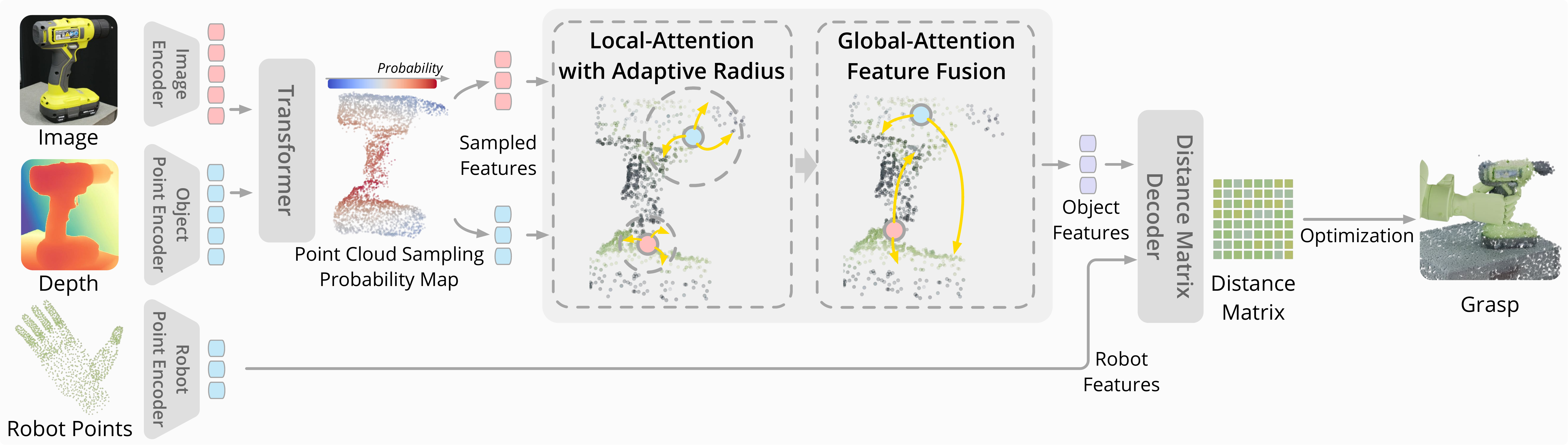}
    \vspace{-0.7cm}
    \caption{
    \textbf{CorDex grasp prediction network.} 
    The network integrates semantic and geometric information from single-view RGB-D input to predict functional dexterous grasps for novel objects. 
    Image and point cloud features are first encoded into pointwise features and processed by a transformer. 
    To boost performance and computational efficiency, we introduce an importance-aware sampling mechanism that samples points around contact areas. 
    Given the sampled points, a local–global fusion module refines local details and encodes holistic object context through global attention. 
    Finally, a distance matrix between the robot hand and object points is decoded via cross-attention and optimized to obtain the final grasp.
    }
    \vspace{-0.5cm}
    \label{fig:policy}
\end{figure*}

\section{Grasp Prediction via Multimodal Fusion}
\label{subsec:policy}

We propose a novel functional dexterous grasp prediction model trained on the large-scale dataset generated by our data pipeline. The model builds on the $\mathcal{D(R,O)}$ representation~\cite{Wei2024DROG} introduced in Sec.~\ref {sec:preliminaries}, where a grasp is encoded as a dense distance matrix between sampled points on the robot hand and the object, and the final grasp $g$ is recovered through multilateration and inverse kinematics. Unlike $\mathcal{D(R,O)}$, which relies exclusively on point cloud geometry, our approach jointly exploits semantic cues from RGB images and geometric properties from depth observations. This multimodal integration enables the model to predict grasps that simultaneously satisfy \emph{stability} and \emph{functionality}, both of which are essential for dexterous functional grasping.

As illustrated in Fig.~\ref{fig:policy}, the network takes as input a single-view RGB-D observation of the object together with the robot hand point cloud. The depth channel is converted into a 3D point cloud, while semantic features are extracted from the RGB image and back-projected onto the same 3D points. Each point is thus represented by both geometric and semantic embeddings computed with image and point cloud encoders~\cite{Oquab2023DINOv2LR, Wang2018DynamicGC}. These multimodal pointwise features are fused into a unified object representation, which is cross-attended with the robot hand features to predict the distance matrix. Two design challenges motivate our architecture: (i) functional regions such as triggers or buttons are often small and easily missed by uniform point sampling, and (ii) accurate grasping requires reasoning at both local detail and global object context. To address these challenges, we introduce two central components: an \emph{importance-aware sampling module} that adaptively focuses on contact-relevant regions, and a \emph{local-global fusion module} that integrates complementary semantic and geometric cues into a coherent representation.

\textbf{Local-global fusion of multimodal features.}
Functional dexterous grasping requires reasoning over both the fine-grained geometry of contact regions and the broader semantic context of the object. 
To address this, we design a local-global fusion module that adaptively combines semantic and geometric information with different receptive fields.
Specifically, semantic features from the RGB image are back-projected onto the sampled 3D points and paired with their geometric features. A local cross-attention mechanism then relates geometric features to nearby semantic features, and vice versa, enabling the network to capture contact-relevant detail. To account for varying point densities, we introduce an \emph{adaptive attention radius}, which uses a larger receptive field to gather richer context in sparse regions, while focusing more narrowly for precision in dense regions. The resulting locally fused features are further processed with global self-attention to encode the object’s overall structure. Finally, local and global features are integrated into a unified representation, which is cross-attended with robot hand features to predict the distance matrix as described in Sec.~\ref {sec:preliminaries}.

\textbf{Efficient prediction via adaptive sampling.}
Encoding all object points uniformly is both computationally inefficient and ineffective for functional grasping, since small functional regions (\textit{e.g.}, triggers, buttons) may be overwhelmed by irrelevant surface points. To focus computation on task-relevant areas, we introduce an importance-aware sampling module that adaptively preserves points likely to lie in contact regions. 
Given the concatenated semantic and geometric features of each object point, a lightweight transformer estimates pointwise importance probabilities using global self-attention to incorporate object context. Guided by this distribution, the point cloud is downsampled from $N=4096$ to $N'=1024$ points, increasing density in functional regions while reducing redundancy elsewhere. Ground-truth importance maps are derived from distances between object points and ground-truth robot hand points, and the sampling module is trained with the KL divergence to match this distribution. This adaptive sampling not only improves computational efficiency but also enhances the model’s ability to capture the subtle object regions essential for functional grasp prediction.

\begin{table*}[t]
\centering
\vspace{-0.3cm}
\caption{\textbf{Quantitative comparisons in simulation.} We evaluate our method against state-of-the-art approaches on all nine tasks using two robotic hands: the 22-DoF Shadow Hand and the 6-DoF Inspire Hand. Reported success rates (\%) indicate grasps that satisfy both stability and functionality. The best results are shown in \textbf{bold}. * denotes one-shot methods}

\vspace{-1em}
\label{tab:tap_vid}
\begin{adjustbox}{max width=1.0\linewidth, center}
\setlength\tabcolsep{7pt} %
\begin{tabular}{ll|cccccccccc}
\toprule
Embodiment & Method & Drill & Pipette & Stapler & Spray Bottle & Hammer & Syringe & Hair Dryer & Aerosol Can & Glue Gun & Avg. \\
\midrule
\multirow{6}{*}{\textit{Shadow}} 
& \emph{D(R,O)}~\cite{Wei2024DROG} & 24.0 & 11.7 & 23.3 & 19.0 & 14.3 & 28.0 & 8.7 & 25.3 & 10.0 & 18.3\\
& \emph{D(R,O)}~\cite{Wei2024DROG}  with our data & 37.7 & 20.7 & 33.7 & 37.7 & 21.0 & 48.0 & 70.0 & 33.3 & 21.7 & 36.0\\
& SparseDFF*~\cite{Wang2023SparseDFFSF} & 7.7 & 14.7 & 15.7 & 16.3 & 22.0 & 18.7 & 11.3 & 17.0 & 9.7 & 14.8\\
& DenseMatcher*~\cite{Zhu2024DenseMatcherL3} & 14.3 & 16.7 & 15.3 & 19.7 & 25.3 & 16.3 & 18.3 & 15.0 & 11.0 & 16.9\\
& AG-Pose~\cite{Lin2024InstanceAdaptiveAG}  with our data& 67.7 & 65.3 & 76.0 & 63.3 & 77.0 & 71.3 & 69.0 & 59.0 & 58.7 & 67.5\\
& \textbf{Ours} & \textbf{90.0} & \textbf{91.3} & \textbf{85.0} & \textbf{85.3} & \textbf{85.7} & \textbf{91.7} & \textbf{98.7} & \textbf{84.3} & \textbf{84.7} & \textbf{88.5} \\
\midrule
\multirow{5}{*}{\textit{Inspire}}
& \emph{D(R,O)}~\cite{Wei2024DROG} with our data & 13.3 & 11.7 & 17.0 & 26.3 & 18.0 & 13.7 & 25.0 & 7.7 & 25.3 & 17.6\\
& SparseDFF*~\cite{Wang2023SparseDFFSF} & 8.0 & 6.0 & 10.3 & 2.3 & 12.3 & 7.3 & 9.7 & 6.3 & 7.7 & 7.8\\
& DenseMatcher*~\cite{Zhu2024DenseMatcherL3} & 5.3 & 6.3 & 7.0 & 4.7 & 14.3 & 7.7 &  12.3& 7.0 & 3.7 & 7.6\\
& AG-Pose~\cite{Lin2024InstanceAdaptiveAG} with our data & 41.3 & 47.0 & 60.7 & 56.0 & 58.3 & 49.0 & 40.0 & 44.3 & 43.3 & 48.9 \\
& \textbf{Ours} & \textbf{72.7} & \textbf{63.3} & \textbf{80.3} & \textbf{87.7} & \textbf{78.0} & \textbf{73.0} & \textbf{75.3} & \textbf{70.7} & \textbf{71.0} & \textbf{74.7}\\
\bottomrule
\vspace{-3.5em}
\label{simresults}
\end{tabular}
\end{adjustbox}
\end{table*}

\section{Experiments}
\label{sec:experiment}
We conduct extensive experiments in both simulation and the real world to evaluate the effectiveness of our approach. Specifically, we aim to answer three key questions:
1) Does the proposed \MethodAcronym data engine generate diverse and high-quality datasets for functional dexterous grasping?
2) Can the \MethodAcronym prediction model effectively infer functional grasps for novel object instances and categories from single-view RGB-D input?
3) What are the critical design factors that contribute to the performance of our model?

\begin{figure}[t]
    \centering
    \includegraphics[width=1.0\linewidth]{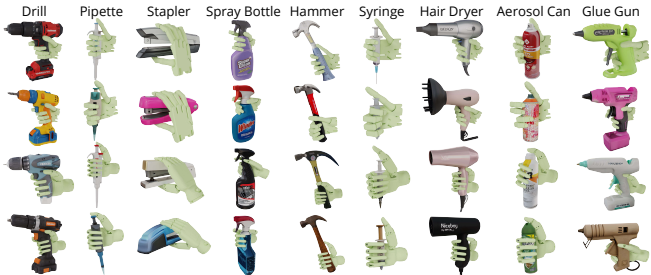}
    \vspace{-0.8cm}
    \caption{\textbf{Examples of generated data.} We generate a functional dexterous grasp dataset consisting of 900 objects, 1.08 million images,
    and 11 million image–grasp pairs. The dataset spans across nine tasks and two different embodiments of different DoFs (Shadow and Inspire). 
    }
    \vspace{-0.6cm}
    \label{fig:taskexamples}
\end{figure}

\subsection{Experimental Setup}
\label{subsec:setup}
\noindent \textbf{Evaluation environments and protocols.}
We consider both the stability and functionality of predicted grasps using simulation and real-world experiments, reporting the success rate of grasps satisfying both requirements.

\emph{Simulation.} 
We conduct experiments across nine object categories using a held-out test set with object models, annotated functional regions, and avoidance regions that should not be touched by the hand. Generated grasps are validated in IsaacGym~\cite{Makoviychuk2021IsaacGH} under external forces on two embodiments: the 22-DoF Shadow Hand and the 6-DoF Inspire Hand, following the protocol in~\cite{Wei2024DROG}. A grasp is considered stable if object displacement is $<2$ cm after external forces are applied, while it is considered functional if the distance between robot hand and functional region $<1$ mm and no avoidance regions (e.g. drill head) are touched.

\begin{figure*}[t]
    \centering
    \vspace{-0.5cm}
    \includegraphics[width=1.0\linewidth]{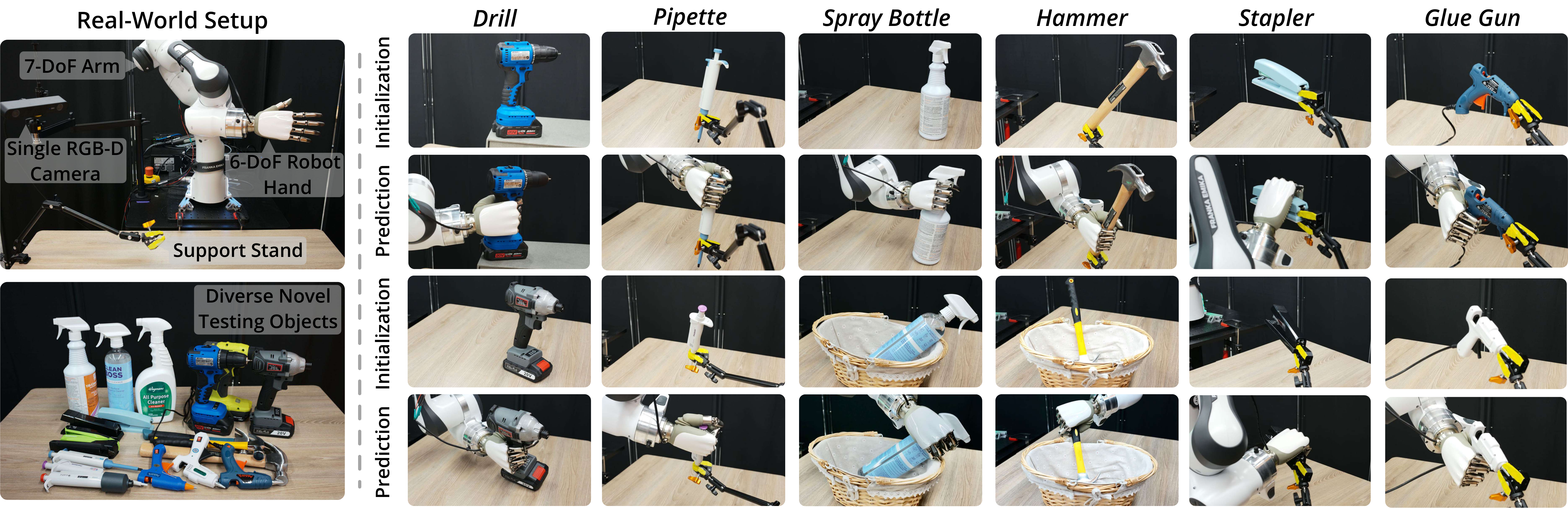}
    \vspace{-0.8cm}
    \caption{
    \textbf{Real-world experiments.} 
    Left: a 7-DoF robot arm with a 6-DoF dexterous hand executes functional grasps predicted by our model from single-view RGB-D input. 
    We evaluate six tasks, each with three real-world objects that are unseen in the generated dataset. 
    Right: qualitative results on these objects, demonstrating category-level generalization to diverse shapes and varying poses.
    }
    \vspace{-0.5cm}
    \label{fig:qualtative}
\end{figure*}

\emph{Real-world.} We evaluate across six object categories, each containing 3 objects, using the 6-DoF OYMotion hand mounted on a 7-DoF Franka Research 3 arm, as shown in Fig.~\ref{fig:qualtative}.
The OYMotion hand is nearly identical to the Inspire Hand but is produced by a different manufacturer.
A ZED camera is mounted on either side of the table and provides single-view \mbox{RGB-D} input. 
We use Grounded SAM~\cite{liu2023grounding,kirillov2023segany} to segment the target object. 
We evaluate on five poses of each object.

\vspace*{1mm}

\vspace*{1mm}
\noindent \textbf{Baseline methods.}
Our method is compared with three categories of approaches:
(1) Dexterous hand grasp prediction method: $\mathcal{D(R,O)}$\cite{Wei2024DROG}, which directly predicts grasps. We also report the results of $\mathcal{D(R,O)}$ trained on our dataset.
(2) One-shot correspondence methods: SparseDFF~\cite{Wang2023SparseDFFSF} and DenseMatcher~\cite{Zhu2024DenseMatcherL3}. Since these methods cannot handle single-view input, we provide two-view RGB-D images in real-world experiments (captured from different viewpoints for a more complete observation) and complete object models in simulation.
(3) Category-level object pose estimation method: AG-Pose~\cite{Lin2024InstanceAdaptiveAG}, which is trained on our dataset for each category to ensure fair comparison.

\subsection{\MethodAcronym Dataset}
\label{subsec:dataset}
Our data engine generates high-quality functional grasp data for diverse objects covering nine common functional object categories, enabling the grasp prediction model to robustly generalize to unseen objects. The object categories include: \textit{Drill, Pipette, Stapler, Spray Bottle, Hammer, Syringe, Hair Dryer, Aerosol Can}, and \textit{Glue Gun}, as shown in Fig.~\ref{fig:taskexamples}. For each category, we create 100 diverse objects with varying shapes and appearances but consistent functionality. For each object, 10 valid functional grasps are generated for both the Shadow Hand and Inspire Hand, two widely used robotic embodiments.
We render 
1,200 RGB-D images per object under diverse poses and lighting conditions. In total, the dataset contains 900 objects, 1.08 million images, and around 11 million image–grasp pairs, generated in $\sim$3 days using 48 NVIDIA A100 GPUs. For each task, 2 objects are held out for validation and 3 for testing, ensuring that evaluation is performed on unseen instances.

\subsection{Comparative Results}
\label{subsec:results}
The quantitative results of simulation and real-world experiments are reported in Tab.~\ref{simresults} and Tab.~\ref{realresults}, while qualitative results are shown in Fig.~\ref{fig:qualtative}.
Compared with \emph{grasp prediction method} $\mathcal{D(R,O)}$~\cite{Wei2024DROG}, our method achieves significantly higher performance in both simulation and real-world settings.
$\mathcal{D(R,O)}$ performs poorly due to the lack of functional grasp supervision in its original training data; retraining it on our dataset improves results but still performs substantially worse than our approach, highlighting the effectiveness of our model design in utilizing visual and features.
Compared with \emph{one-shot correspondence} methods SparseDFF~\cite{Wang2023SparseDFFSF} and DenseMatcher~\cite{Zhu2024DenseMatcherL3}, which are provided with more complete observations, our single-view method also outperforms them by a large margin. These methods suffer from the lack of generalizability of dense 3D correspondence, which is difficult to learn from limited correspondence data, leading to inaccurate contact transfer and low-quality grasps.
Finally, compared with the \emph{category-level pose estimation} method AG-Pose~\cite{Lin2024InstanceAdaptiveAG}, our approach achieves substantially better results. Even when trained on our dataset, category-level methods rely on coarse object alignment, which is insufficient for functional grasping that requires precise contact with functional regions. In contrast, our method directly predicts grasp gestures without explicit alignment, demonstrating generalization to unseen objects with diverse shapes.
\begin{table}[t]
\centering
\caption{\textbf{Quantitative comparison in the real-world.} Success counts for functional grasping across six tasks. Best results are shown in \textbf{bold}. * denotes one-shot methods.}
\vspace{-1em}
\label{tab:realworld}
\begin{adjustbox}{max width=\linewidth, center}
\setlength\tabcolsep{2.5pt} %
\begin{tabular}{lccccccc}
\toprule
&
\multicolumn{1}{c}{Drill} & 
\multicolumn{1}{c}{Pipette} & 
\multicolumn{1}{c}{Stapler} & 
\multicolumn{1}{c}{Spray Bottle} & 
\multicolumn{1}{c}{Hammer} & 
\multicolumn{1}{c}{Glue Gun} \\
\midrule
\emph{D(R,O)}~\cite{Wei2024DROG} with our data & 2/15 & 0/15 & 3/15 & 2/15 & 4/15 & 2/15\\
SparseDFF*~\cite{Wang2023SparseDFFSF} & 3/15 & 0/15 & 3/15 & 1/15 & 3/15 & 1/15\\
DenseMatcher*~\cite{Zhu2024DenseMatcherL3} & 1/15 & 0/15 & 2/15 & 0/15 & 3/15 & 0/15\\
AG-Pose~\cite{Lin2024InstanceAdaptiveAG} with our data& 3/15 & 2/15 & 6/15 & 3/15 & 9/15 & 4/15\\
\textbf{Ours} & \textbf{10}/\textbf{15} & \textbf{7}/\textbf{15} & \textbf{11}/\textbf{15} & \textbf{11}/\textbf{15} & \textbf{13}/\textbf{15} & \textbf{10}/\textbf{15}\\ 
\bottomrule
\vspace{-4.5em}
\label{realresults}
\end{tabular}
\end{adjustbox}
\end{table}

\vspace*{1mm}
\noindent \textbf{Running time.}
The end-to-end inference time per observation, including distance matrix prediction and grasp optimization, is 0.92 s on the Shadow Hand and 0.36 s on the low-DoF Inspire Hand, measured on an NVIDIA 4090 GPU.

\subsection{Ablation Studies}
\label{subsec:ablation}
We conduct ablation studies on six tasks in simulation with the Inspire Hand to evaluate the contribution of each design component in both the data generation pipeline and the model.
For the \emph{data generation}, we first replace our correspondence transfer with a 3D matching method~\cite{Zhu2024DenseMatcherL3}. As shown in Tab.~\ref{tab:ablation}(1), this substitution leads to a significant performance drop due to inaccurate contact transfer and low-quality grasp optimization, highlighting the effectiveness and flexibility of our robust correspondence transfer approach. We further ablate the design of preserving multiple transferred contact points during optimization (Tab.~\ref{tab:ablation}(2)). Retaining only a single contact point results in degraded performance, confirming the importance of multiple candidates for handling transfer noise.
For the \emph{model architecture}, we first remove the image input and use only point clouds (Tab.~\ref{tab:ablation}(3)). This causes a large performance drop, as single-view point clouds provide ambiguous semantics, while our model benefits from complementary image features. Next, we ablate the importance-aware sampling (Tab.~\ref{tab:ablation}(4)). Without it, fewer points near the hand are preserved and performance decreases noticeably. Finally, removing the local attention module with adaptive radius (Tab.~\ref{tab:ablation}(5)) consistently lowers performance, demonstrating its effectiveness in capturing fine-grained local context as a complement to global attention.
\begin{table}[t]
\centering
\caption{\textbf{Ablation studies in simulation.} We report success rate (\%) for each variant. best results are shown in \textbf{bold}.}
\vspace{-1.0em}
\begin{adjustbox}{max width=\linewidth, center}
\setlength\tabcolsep{1.5pt} %
\begin{tabular}{lccccccc}
\toprule
Method & Drill & Pipette & Stapler  & Hammer & Aerosol Can & Glue Gun & Avg. \\
\midrule
Full & \textbf{72.7} & \textbf{63.3} & \textbf{80.3} & \textbf{78.0} & \textbf{70.7} & \textbf{71.0} & \textbf{72.7}\\
\midrule
(1) Data engine with 3D matching~\cite{Zhu2024DenseMatcherL3} & 25.3 &  14.0 & 16.3 & 11.7 & 18.7 & 25.0 & 18.5\\
(2) Data engine \textit{w/o} multiple candidates & 67.3 & 59.7 & 71.0 & 71.3 & 62.0 & 65.3 & 66.1\\
\midrule
(3) Grasp network \textit{w/o} image input & 20.0 & 18.0 & 23.3 & 15.0 & 19.7 & 28.0 & 20.7\\
(4) Grasp network \textit{w/o} importance sampling & 64.7 & 57.0 & 73.3 & 66.3& 60.7 & 68.3 & 65.1\\
(5) Grasp network \textit{w/o} local attention & 47.3 & 51.7 & 60.0 & 57.0 & 51.3 & 49.0 & 52.7\\
\bottomrule
\vspace{-4.5em}
\label{tab:ablation}
\end{tabular}
\end{adjustbox}
\end{table}

\section{Conclusion}
In this paper, we present a novel framework for dexterous functional grasping that integrates a correspondence-based data engine with a grasp prediction network employing local-global adaptive feature fusion. The data engine autonomously generates large-scale functional grasp data for diverse objects from a single human demonstration, while the prediction network effectively leverages both visual and geometric features to infer accurate functional grasps from single-view RGB-D input. Extensive experiments in both simulation and real-world settings show that our approach substantially outperforms state-of-the-art baselines.
Furthermore, our data engine can be readily extended to new tasks without additional training, offering a scalable pipeline for data curation and paving the way toward universal dexterous grasping models.

\vspace*{1mm}
\noindent \textbf{Limitations.}
Despite these advances, there are two main limitations of the proposed approach. First, although depth noise is injected during training, the model remains sensitive to severely corrupted or displaced depth input in the real world, reflecting the domain gap between synthetic and real-world depth sensing. Second, while the framework generalizes to novel object instances, it still focuses on category-specific training and does not yet handle fully open-set scenarios. Future work should explore scaling task diversity and developing universal models that exhibit emergent generalization to unseen objects and tasks.

{
\small
\bibliographystyle{IEEEtran}
\bibliography{IEEEexample.bib}
}

\end{document}